\documentclass[10pt,twocolumn,letterpaper]{article}

\usepackage[pagenumbers]{iccv} 

\PassOptionsToPackage{table}{xcolor}
\usepackage{xcolor}
\usepackage{iccv}
\usepackage{times}
\usepackage{colortbl}      

\usepackage{amsmath}
\usepackage{amssymb}

\usepackage{verbatim}
\usepackage{makecell}
\usepackage{multirow}
\usepackage{multicol}
\usepackage{float}
\usepackage{placeins}
\usepackage{pifont}       
\usepackage{arydshln}
\usepackage{booktabs}
\usepackage{tabu}
\usepackage[symbol]{footmisc}
\usepackage{soul}
\usepackage{academicons}
\usepackage[normalem]{ulem}
\usepackage{graphicx}
\DeclareUnicodeCharacter{0308}{\"{}}

\definecolor{Gray}{gray}{0.9}
\definecolor{dGray}{gray}{0.6}
\definecolor{lowGray}{gray}{0.3}
\definecolor{OliveGreen}{rgb}{0,0.6,0}
\definecolor{red}{rgb}{0.6,0,0}

\usepackage{placeins}
\usepackage{cuted}      
\usepackage{tocloft}   
\usepackage{lipsum}    

\makeatletter
\newcommand{\listappendices}{%
  \section*{Table of Contents}%
  \@starttoc{app}%
}
\newcommand{\appsection}[1]{%
  \section{#1}%
  \addcontentsline{app}{section}{\protect\numberline{\thesection}#1}%
}
\newcommand{\appsubsection}[1]{%
  \subsection{#1}%
  \addcontentsline{app}{subsection}{\protect\numberline{\thesubsection}#1}%
}
\makeatother

\definecolor{iccvblue}{rgb}{0.21,0.49,0.74}
\usepackage[pagebackref,breaklinks,colorlinks,allcolors=iccvblue]{hyperref}

\def\paperID{457} 
\def\confName{ICCV}
\def\confYear{2025}

\title{Enhancing Retinal Vessel Segmentation Generalization via Layout-Aware Generative Modelling}

\author{
Jonathan Fhima$^{1,2}$ \quad Jan Van Eijgen$^{3,4}$ \quad Lennert Beeckmans$^{4,5}$ \quad Thomas Jacobs$^{4}$\\ 
 Moti Freiman$^{1}$ \quad Luis Filipe Nakayama$^{6,7}$ Ingeborg Stalmans$^{3,4}$ \quad Chaim Baskin$^{8\dagger}$ \quad Joachim A. Behar$^{1\dagger}$ \\
{\tt\small jbehar@technion.ac.il} \\
{\small $^\dagger$Equal contribution as Principal Investigators.} \\
\thanks{
{\fontsize{6}{8}\selectfont
\setlength{\baselineskip}{6pt} 
$^{1}$Faculty of Biomedical Engineering, Technion-IIT, Israel. \\
$^{2}$Faculty of Mathematics, Technion-IIT, Israel. \\
$^{3}$Department of Neurosciences, KU Leuven, Belgium. \\
$^{4}$Department of Ophthalmology, UZ Leuven, Belgium. \\
$^{5}$Department of Electrical Engineering, KU Leuven, Belgium.\\
$^{6}$Ophthalmology Department,São Paulo Federal University, Brazil. \\
$^{7}$Medical Engineering and Science, Massachusetts Institute of Technology, USA. \\
$^{8}$School of Electrical and Computer Engineering, Ben-Gurion University of the Negev, Israel.}
}
}

\newcommand{\indomain}{Local}
\newcommand{\neardomain}{External}
\newcommand{\outdomain}{OOD}
\newcommand{\methodname}{RLAD}
\newcommand{\datasetname}{REYIA}
\newcommand{\component}{layout component}
\newcommand{\cmark}{\ding{51}}
\newcommand{\xmark}{\ding{55}}

\begin{document}
\maketitle

\begin{abstract}
Generalization in medical segmentation models is challenging due to limited annotated datasets and imaging variability. To address this, we propose \textbf{R}etinal \textbf{L}ayout-\textbf{A}ware \textbf{D}iffusion (\methodname{}), a novel diffusion-based framework for generating controllable layout-aware images. \methodname{} conditions image generation on multiple key \component s extracted from real images, ensuring high structural fidelity while enabling diversity in other components. Applied to retinal fundus imaging, we augmented the training datasets by synthesizing paired retinal images and vessel segmentations conditioned on extracted blood vessels from real images, while varying other \component s such as lesions and the optic disc. Experiments demonstrated that \methodname{}-generated data improved generalization in retinal vessel segmentation by up to 8.1\%. Furthermore, we present \datasetname{}, a comprehensive dataset comprising 586 manually segmented retinal images. To foster reproducibility and drive innovation, both our code and dataset will be made publicly accessible (upon publication).
\end{abstract}
\section{Introduction} \label{sec}
\begin{figure}[t] 
  \centering
  \includegraphics[width=\linewidth]{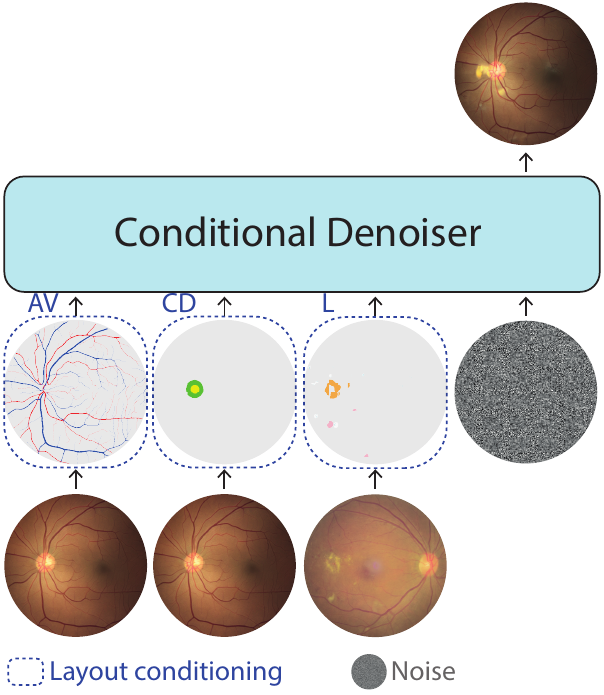}
  \caption{\textbf{Retinal Layout-Aware Diffusion} generates realistic retinal images from noise and user-defined \component s; artery/vein (AV), optic cup/disc (CD), and lesions (L).} 
  \label{fig:teaser}
\end{figure}
\begin{figure*}[ht] 
  \centering
  \includegraphics[width=  \linewidth]{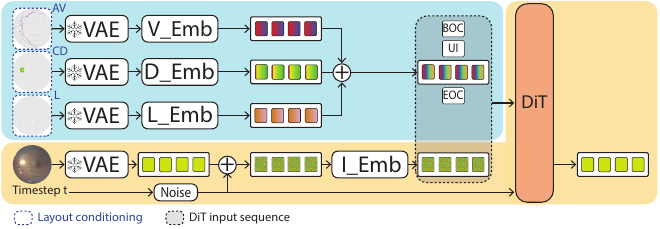}
  \caption{\textbf{\methodname{} Architecture.} The original fundus image and segmentation maps for artery/vein (AV), the optic cup/disc (CD), and lesions (L) are encoded into latent representations using a frozen VAE. Gaussian noise is added to the image latent, and each latent (image, CD, AV, and L) is projected into the DiT \cite{peebles2023scalable} input space via distinct projections. Condition embeddings for AV, CD, and L are summed into a single embedding, \(c\). The DiT input consists of a beginning-of-conditioning (BOC) token, user input (UI), \(c\), an end-of-conditioning (EOC) token, and the noised image latent. The DiT outputs the corresponding denoised image latent. The UI token specifies whether a \component{} is guided by user input or defaults to a neutral embedding when absent.} 
  \label{fig:RAD_arch}
\end{figure*}

Deep learning has achieved remarkable success across various domains, but its progress often depends on access to large annotated datasets. In fields such as natural language processing, vision-language modeling, and image generation, synthetic data from large models has driven significant advancements \cite{wasserman_paint_2024,liu2024visual,li_data_2024,smollm_corpus_dataset,smollm_models_collection,zhu_minigpt-4_2023}. However, in medical imaging, particularly retinal vessel segmentation, data scarcity and variability in imaging conditions remain persistent limitations \cite{galdran2022state,shi_deep_2022,fhima2024lunet,khan2024rvd}. Retinal vessel segmentation is critical for the diagnosis of ocular and systemic diseases \cite{wong_retinal_2003,liew2008retinal,huang_ai-integrated_2024,frost2013retinal}, yet the creation of annotated datasets demands a considerable amount of time, specialized expertise, and consistency across imaging devices \cite{fhima2022lirot}.

Retinal vessel segmentation involves two tasks: general vessel segmentation, which identifies the vasculature, and artery/vein (AV) segmentation, which also differentiates arteries from veins. This distinction provides insights into vessel-specific pathologies\cite{orlando2018towards,fhima2022pvbm}. However, AV segmentation requires complex annotations, making it challenging to obtain sufficient labeled data for robust training.

Generative models like GANs and VAEs have been explored to address data scarcity in medical imaging \cite{goodfellow_generative_2020,kingma2013auto}. When applied to retinal images, these models often encounter challenges, including difficulties in preserving anatomical fidelity and issues with training stability \cite{go2024generation}. Diffusion models have recently emerged as powerful tools for generating diverse high-fidelity images, with superior stability and detail preservation, compared to GANs and VAEs \cite{ho2020denoising,dhariwal2021diffusion}. Despite their success in image synthesis tasks across domains, e.g., natural image generation and text-to-image modeling, their application in medical imaging has largely focused on generating synthetic images rather than directly enhancing segmentation performance through data augmentation.



To address these limitations, we propose Retinal Layout-Aware Diffusion \textbf{(\methodname{})}, a diffusion-based framework for the controllable generation of synthetic retinal images (Figure~\ref{fig:teaser}). By conditioning on multiple key retinal structures—such as artery/vein (AV), the optic cup/disc (CD), and lesions (L)—\methodname{} preserves essential vascular layouts while introducing variability in other regions. This enables the creation of paired image-segmentation maps that expand training datasets without compromising structural integrity. Synthetic data generated by \methodname{} improve segmentation model robustness across diverse imaging conditions and acquisition settings.

We evaluated \methodname{}-generated data using state-of-the-art visual encoders such as Vision Transformers \cite{dosovitskiy2021image} and Swin Transformers \cite{liu2021swin}, and demonstrate consistent improvements in generalization performance under distribution shifts (up to 8.1\%). Additionally, we introduce \datasetname{}, the largest multi-source collection of 586 retinal images with human reference AV segmentation, which not only complements our synthetic data but also demonstrates strong baseline performance, further validating the effectiveness of our synthetic data.

In summary, the main contributions of this work are:
\begin{itemize}
    \item A novel multi-layout-aware generative model (\textbf{\methodname{}}) that synthesizes diverse yet anatomically accurate retinal images while preserving semantic structures.
    \item Demonstrating consistent segmentation performance improvements across state-of-the-art architectures using \methodname{}-generated data.
    \item Introducing \textbf{\datasetname{}}, the largest multi-source collection of datasets for AV-segmented retinal fundus images.
\end{itemize}

\section{Related Work}
Retinal AV segmentation plays a critical role in diagnosing microvascular pathologies \cite{Gunn1892OphthalmoscopicTension,Scheie1953EvaluationSclerosis,Keith1939SomePrognosis,Sharrett1999RetinalStudy,Witt2006AbnormalitiesStroke}. Early methods \cite{Hemelings2019ArteryveinNetwork,Hu2021AutomaticImages,Zhou2021LearningImaging,zhou_automorph_2022,shi_deep_2022}, such as Little W-Net \cite{galdran2022state}, focused on compact convolutional neural networks to reduce computational complexity. More recently, LUNet achieved state-of-the-art performance on optic disc-centered images but struggled to generalize to macula-centered images \cite{fhima2024lunet}. This underscores the primary challenge of achieving robust generalization across diverse retinal imaging conditions.


Generative adversarial networks have been extensively used for retinal image synthesis, often conditioning the generation process on features such as vessel or lesion masks \cite{costa_end--end_2018, zhao_synthesizing_2018}. While these methods produced visually realistic images, they frequently lacked anatomical accuracy and robustness \cite{go2024generation}, limiting their effectiveness for downstream tasks like AV segmentation. To address these issues, Go et al. \cite{go2024generation} proposed a hybrid approach that combined a diffusion model for generating AV masks with a conditional GAN for synthesizing retinal images. Their method preserved patient privacy and demonstrated that synthetic images could lead to AV segmentation performance comparable to models trained on real data. However, it failed to further enhance AV segmentation performance further, possibly due to limited variability in the generated AV masks, which may have propagated to the synthesized images.

Diffusion models have demonstrated remarkable generative capabilities across various domains, including image synthesis, video generation, layout and 3D modeling \cite{sohl2015deep,ho2020denoising,kong2020diffwave, saharia2022palette,ho2022video,poole_dreamfusion_2022,levi2023dlt,  scotti2024reconstructing,wasserman_paint_2024}. Recent advancements, such as classifier-free guidance \cite{ho2021classifierfree} enable precise control over conditioning signals during generation, making these models well-suited for structured image synthesis tasks. Transformer-based architectures such as DiT \cite{peebles2023scalable} further enhance performance by capturing long-range dependencies.

Building on these developments, we propose a multi-layout-aware diffusion framework specifically designed for retinal fundus image synthesis. Unlike prior approaches, our method conditions generation on multiple retinal \component s —AV, CD, and L—extracted from real, non-annotated images using pretrained segmentation models. This minimizes error propagation and enhances realism while addressing domain generalization challenges in AV segmentation tasks through synthetic data augmentation.

\section{Datasets}
This section introduces the new datasets created for this study and provides an overview of the datasets used for diffusion model training and downstream segmentation tasks. For additional details, please refer to the supplementary material.

\subsection{New Datasets}
We introduce \datasetname{}, a curated set of 586 retinal fundus images annotated with AV blood vessel segmentations using the open-access Lirot.ai software \cite{fhima2022lirot}. To enhance diversity, \datasetname{} includes manually segmented images as part of this research from nine datasets: FIVES \cite{jin2022fives}, TREND \cite{popovic_trend_2021}, GRAPE \cite{huang2023grape}, MESSIDOR \cite{messidor2011}, MAGRABIA \cite{Almazroa2018RIGA}, PAPILA \cite{kovalyk2022papila}, MBRSET \cite{wu2024mbrset} AV-WIDE \cite{estrada2015retinal} and ENRICH. ENRICH is a new dataset collected for this study, consisting of 111 retinal fundus images. AV-WIDE, which initially contained only skeletonized vessels, was reannotated to include complete vessel segmentations.

\subsection{Diffusion Model Datasets}
To train \methodname{}, we curated 112,320 retinal fundus images from publicly available datasets spanning diverse imaging conditions, fields of view (FOV), and pathologies. The sources include widely used datasets: UZLF ~\cite{van_eijgen_leuven-haifa_2024}, GRAPE~\cite{huang2023grape}, MESSIDOR~\cite{messidor2011}, PAPILA~\cite{kovalyk2022papila}, MAGRABIA~\cite{Almazroa2018RIGA}, ENRICH, 1000images~\cite{cen_automatic_2021}, DDR~\cite{DDR2019}, EYEPACS~\cite{EYEPACS}, G1020~\cite{G1020}, IDRID~\cite{IDRID2018} and ODIR~\cite{ODIR2019}. Evaluation of the realism of the generated images, in comparison to real images, was performed on the DRTiD dataset \cite{hou2022cross}.

\begin{figure*}[ht] 
  \centering
    \includegraphics[width=\linewidth]{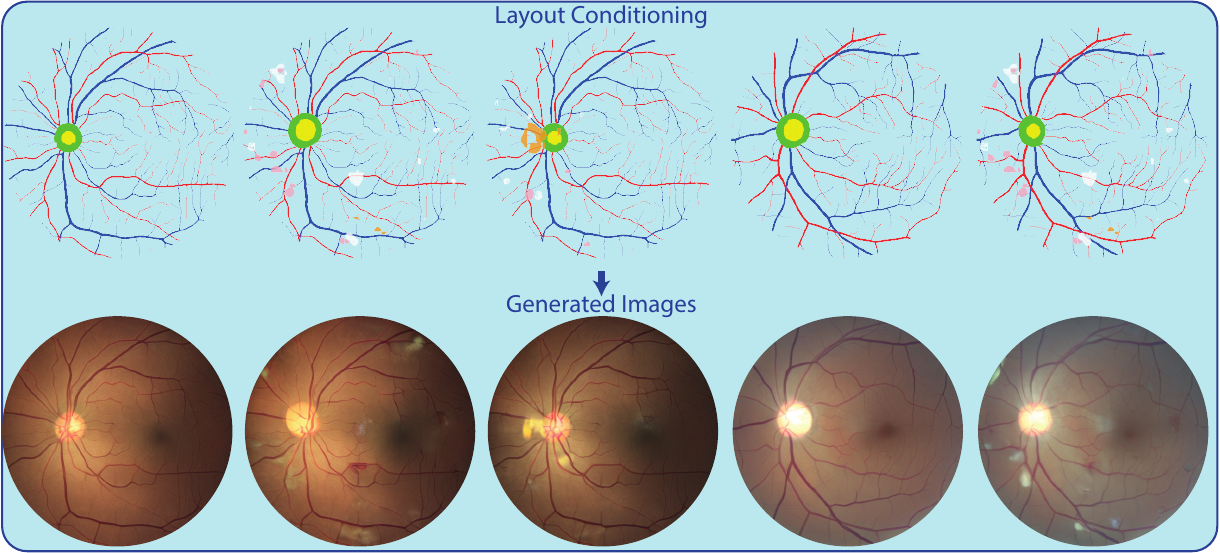}
  \caption{\textbf{Retinal Layout-Aware Diffusion Qualitative Examples.} Top: user-defined \component s inputs (artery/vein in red/blue, optic disc/cup in green/yellow, and lesions in white/pink/orange). Bottom: corresponding generated fundus images.}

    \label{fig:generation_examples}
\end{figure*}

\subsection{AV Segmentation Datasets}
  \label{sec:segmentation_datasets}

\subsubsection{Datasets for Segmentation Model Training}
To train our segmentation models, we constructed a composite dataset combining the UZLF dataset with newly annotated versions of GRAPE, MESSIDOR, ENRICH, MAGRABIA, and PAPILA. These datasets feature high-resolution retinal fundus images with FOVs ranging from 30° to 45° and encompass a variety of ophthalmic conditions and patient populations.

\subsubsection{Datasets for Segmentation Model Evaluation}
To assess generalization performance under varying levels of distribution shift, we evaluated our segmentation models across three categories of datasets:

\textbf{In-Domain (\indomain{}):} Data collected from the same hospital under similar acquisition conditions to those as one of the training datasets, ensuring minimal distribution shifts.

\textbf{Near-Domain (\neardomain{}):} Data from different hospitals and environment, introducing moderate distribution shifts. This category includes HRF~\cite{Budai2013Robust}, INSPIRE~\cite{Niemeijer2011Automated,fhima2024lunet}, UNAF~\cite{Benitez2021Dataset,fhima2024lunet} and the reannotated FIVES dataset.

\textbf{Out-of-Domain (\outdomain{}):} Data that significantly differ from the training distribution, used to evaluate the model robustness across diverse imaging conditions. It includes AV-WIDE for ultra-wide-angle images, IOSTAR~\cite{iostar} for laser-based images, DRIVE~\cite{hu_automated_2013,staal_ridge-based_2004} for low-resolution images, RVD~\cite{khan2024rvd} for video frames from handheld devices, TREND and MBRSET for handheld device images. 

\section{Method}
Our objective is to generate realistic retinal images based on key retinal \component s, specifically AV, CD, and L, extracted from real retinal fundus images.

\subsection{Layout Extraction} 
We extract retinal layouts using open-source models for L segmentation \cite{men2023drstagenet} and CD segmentation \cite{fhima2022pvbm,fhima2024computerized}. For AV segmentation, we retrained a SwinV2$_{\text{tiny}}$-based model on our annotated datasets with data augmentation techniques such as random color jitter, flips, and rotations. These extracted retinal \component s serve as input to the diffusion process.
The impact of the layout extractor used is further discussed in the supplementary material.


\subsection{Retinal Layout-Aware Diffusion}
Our approach builds upon latent diffusion \cite{rombach2022high} and DiT \cite{peebles2023scalable}. The forward diffusion process \cite{sohl2015deep,ho2020denoising} gradually adds Gaussian noise to an image $x_0$, producing $x_t$. This process is defined as:
\begin{equation}
q(x_t \mid x_0) = \mathcal{N}(x_t; \sqrt{\overline{\alpha}_t} x_0, (1-\overline{\alpha}_t) I),
\end{equation}
where the noise schedule $\{\overline{\alpha}_t\}$ follows a linear strategy as explored in \cite{ho2020denoising}. The reverse process approximates the denoising steps to reconstruct $x_0$:
\begin{equation}
p_\theta(x_{t-1} \mid x_t, c) = \mathcal{N}(x_{t-1}; \mu_\theta(x_t, c), \Sigma_\theta(x_t, c)),
\end{equation}
where $c$ denotes conditioning information. Instead of operating directly in pixel space, we adopt latent diffusion and perform these operations in a compressed latent space of a frozen VAE. This allows us to refine latent representations $z_t$ iteratively towards $z_0$, improving computational efficiency and scalability.

To incorporate conditional information into the diffusion process, we extract the \component s (\(\mathrm{AV}, \mathrm{CD}\)  and \(\mathrm{L}\)) from the input data. These components are embedded into the transformer's latent space using dedicated projection heads: \(V_{\text{emb}}, D_{\text{emb}}\) and \(L_{\text{emb}}\). 
\[
c_{\text{AV}} = V_{\text{emb}}\bigl(\mathrm{AV}\bigr),
\quad
c_{\text{CD}} = D_{\text{emb}}\bigl(\mathrm{CD}\bigr), 
\quad
c_{\text{L}} = L_{\text{emb}}\bigl(\mathrm{L}\bigr).
\]
To handle both fully and partially conditional cases, we used user input (UI) tokens. Each token indicates whether a component is user-defined (guided) or neutral (unconditional). During training, each layout component is either provided or masked with a certain probability, allowing the model to learn both conditional and unconditional scenarios. This probabilistic masking is applied independently to each component. When a component is masked, it is replaced with a ``black`` image embedding, and its corresponding UI token is updated to signal the absence of guidance:
\[
\text{UI} = [\text{UI}_{\text{AV}},\, \text{UI}_{\text{CD}},\,\text{UI}_{\text{L}}],
\]
allowing flexible control over the conditioning process.
The final conditioning vector is computed as:
\[
c = c_{\text{AV}} + c_{\text{CD}} + c_{\text{L}}.
\]

which is fed into the transformer as part of a sequence:
\[
[\mathrm{BOC},\, \text{UI},\, c,\, \mathrm{EOC},\, z_t],
\]
where \(\mathrm{BOC}\) and \(\mathrm{EOC}\) mark the beginning and end of the conditioning tokens, respectively. After the transformer processes this sequence, only the image tokens are retained to produce \(z_{t-1}\). 
This design ensures that conditioning signals guide the denoising process without remaining entangled in the final latent representation. A schematic overview of our architecture is provided in Figure~\ref{fig:RAD_arch}.

\paragraph{Training Objective.}  
Following DDPM \cite{ho2020denoising}, we adopt a noise prediction loss. Instead of directly modeling $\mu_\theta$ and $\Sigma_\theta$, our model predicts the noise $\epsilon$ added at a randomly chosen timestep $t$:
\begin{equation}
L_{\text{simple}} = \mathbb{E}_{z_0, t, \epsilon}\bigl[ \lVert \epsilon - \hat{\epsilon}_\theta(z_t, t, c) \rVert^2 \bigr].
\end{equation}

Minimizing this MSE loss enables the model to accurately denoise latent representations, effectively learning to reverse the diffusion process. By incorporating tokens that differentiate between user-defined and neutral embeddings for each \component{}, the model can both generate anatomically guided images when specific conditions are provided, and produce diverse, unconstrained samples in the absence of such guidance. This flexibility ensures that the model adapts seamlessly to varying levels of conditional input, balancing anatomical fidelity with generative diversity. 

\textbf{Sampling.}  
To generate new images, we start from a random Gaussian latent $z_T \sim \mathcal{N}(0, I)$ and iteratively remove noise at each diffusion step $t$. Our model predicts the added noise \(\hat{\epsilon}_\theta(z_t, t, c)\), where $c$ includes tokens for AV, CD, and L layouts.

We employ classifier-free guidance \cite{ho2021classifierfree} to control how closely the model adheres to provided conditions. At each step, two predictions are made: one conditional ($c$) and one unconditional ($c=\emptyset$). These are combined as:
\begin{equation}
\hat{\epsilon}_\theta^{\text{guided}}(z_t,t,c) = \hat{\epsilon}_\theta(z_t,t,\emptyset) + w \bigl(\hat{\epsilon}_\theta(z_t,t,c) - \hat{\epsilon}_\theta(z_t,t,\emptyset)\bigr),
\end{equation}
where $w$ is a guidance scale. Higher $w$ yields more faithful adherence to the conditions, lower $w$ allows more diversity.

By iteratively applying guided noise predictions until reaching  $z_0$, we decode $z_0$ using the VAE to produces a synthetic retinal fundus image. This approach balances anatomical fidelity when conditions are provided with greater diversity when they are neutral or absent. Examples of generated images are shown in Figure~\ref{fig:generation_examples}.

\begin{table*}[t!]
  \centering
    \setlength{\tabcolsep}{3pt} 
  \resizebox{\linewidth}{!}{%
  \begin{tabular}{l c ccccc c ccccccc cc}
    \toprule
     \multirow{2}{*}{\textbf{Backbone}} & \multicolumn{2}{c}{\textbf{\indomain{}}} & \multicolumn{4}{c}{\textbf{\neardomain{}}} & \multicolumn{6}{c}{\textbf{\outdomain}} & \multicolumn{2}{c}{\textbf{Average}} \\ 
     & UZLF & LES-AV & HRF & INSPIRE & FIVES & UNAF & AV-WIDE & IOSTAR & DRIVE & RVD & TREND & MBRSET & \neardomain{} & \outdomain{} \\
    \cmidrule(lr){2-3} \cmidrule(lr){4-7} \cmidrule(lr){8-13} \cmidrule(lr){14-15}

    



\selectfont\color{lowGray} RMHAS\cite{shi_deep_2022}
    & \selectfont\color{lowGray} -
    & \selectfont\color{lowGray} 60.0
    & \selectfont\color{lowGray} 48.0
    & \selectfont\color{lowGray} -
    & \selectfont\color{lowGray} -
    & \selectfont\color{lowGray} -
    & \selectfont\color{lowGray} -
    & \selectfont\color{lowGray} 55.0
    & \selectfont\color{lowGray} 60.0
    & \selectfont\color{lowGray} -
    & \selectfont\color{lowGray} -
    & \selectfont\color{lowGray} -
    & \selectfont\color{lowGray} -
    & \selectfont\color{lowGray} - \\

    \selectfont\color{lowGray} RVD$_{\text{Swin-L}}$  \cite{khan2024rvd}
    & \selectfont\color{lowGray} -
    & \selectfont\color{lowGray} -
    & \selectfont\color{lowGray} -
    & \selectfont\color{lowGray} -
    & \selectfont\color{lowGray} -
    & \selectfont\color{lowGray} -
    & \selectfont\color{lowGray} -
    & \selectfont\color{lowGray} -
    & \selectfont\color{lowGray} 57.3 
    & \selectfont\color{lowGray} 53.0
    & \selectfont\color{lowGray} -
    & \selectfont\color{lowGray} -
    & \selectfont\color{lowGray} -
    & \selectfont\color{lowGray} - \\

    \selectfont\color{lowGray} Little W-Net \cite{galdran2022state}
    & \selectfont\color{lowGray} 80.7
    & \selectfont\color{lowGray} 82.0
    & \selectfont\color{lowGray} 58.1
    & \selectfont\color{lowGray} 71.3
    & \selectfont\color{lowGray} 73.5
    & \selectfont\color{lowGray} 68.6
    & \selectfont\color{lowGray} 43.1
    & \selectfont\color{lowGray} 29.9
    & \selectfont\color{lowGray} 61.3 
    & \selectfont\color{lowGray} 34.7
    & \selectfont\color{lowGray} 53.4
    & \selectfont\color{lowGray} 50.4
    & \selectfont\color{lowGray} 67.9
    & \selectfont\color{lowGray} 45.5 \\

    \selectfont\color{lowGray} Automorph \cite{zhou_automorph_2022}
    & \selectfont\color{lowGray} 76.3
    & \selectfont\color{lowGray} 84.0$^{\dagger}$
    & \selectfont\color{lowGray} 77.4$^{\dagger}$
    & \selectfont\color{lowGray} 71.1
    & \selectfont\color{lowGray} 72.5
    & \selectfont\color{lowGray} 65.9
    & \selectfont\color{lowGray} 50.1
    & \selectfont\color{lowGray} 54.9
    & \selectfont\color{lowGray} 78.1$^{\dagger}$
    & \selectfont\color{lowGray} 34.1
    & \selectfont\color{lowGray} 66.6
    & \selectfont\color{lowGray} 63.7
    & \selectfont\color{lowGray} 71.7$^{\dagger}$
    & \selectfont\color{lowGray} 57.9$^{\dagger}$ \\

        \selectfont\color{lowGray} VascX \cite{quiros_vascx_2024}
    & \selectfont\color{lowGray} 80.6
    & \selectfont\color{lowGray} 81.8
    & \selectfont\color{lowGray} 75.6
    & \selectfont\color{lowGray} 74.9
    & \selectfont\color{lowGray} 80.4
    & \selectfont\color{lowGray} 73.1
    & \selectfont\color{lowGray} 49.8
    & \selectfont\color{lowGray} 52.1
    & \selectfont\color{lowGray} 73.6
    & \selectfont\color{lowGray} 42.6
    & \selectfont\color{lowGray} 71.9
    & \selectfont\color{lowGray} 73.2
    & \selectfont\color{lowGray} 76.0
    & \selectfont\color{lowGray} 60.5\\

    \selectfont\color{lowGray} LUNet \cite{fhima2024lunet}
    & \selectfont\color{lowGray} 83.2 
    & \selectfont\color{lowGray} 83.5 
    & \selectfont\color{lowGray} 73.1
    & \selectfont\color{lowGray} 75.5 
    & \selectfont\color{lowGray} 86.0 
    & \selectfont\color{lowGray} 74.4  
    & \selectfont\color{lowGray} 69.3
    & \selectfont\color{lowGray} 56.7
    & \selectfont\color{lowGray} 71.1
    & \selectfont\color{lowGray} 35.2
    & \selectfont\color{lowGray} 71.1 
    & \selectfont\color{lowGray} 63.2 
    & \selectfont\color{lowGray} 77.3 
    & \selectfont\color{lowGray} 61.1 \\


    \addlinespace[0.5em] 
    \cdashline{1-15}
    \addlinespace[0.5em] 



    DinoV2$_{\text{small}}$ \cite{oquab2023dinov2}
    &81.6 
    &82.4 
    &74.2 
    &76.6 
    &82.7 
    &72.9
    &59.4 
    &57.2 
    &75.0  
    &45.4  
    &67.1  
    & 79.6 
    &76.6  
    &64.0\\

        \rowcolor[HTML]{C5E6F2} 
    + \methodname{} (Our)   
    & 81.8
    & 82.8 
    & 75.1
    & 77.5
    & 83.6
    & 73.7
    & 58.3
    & 65.3
    & 76.8
    & 46.7
    & 70.8
    & 81.9
    & 77.5
    & 66.6 
    \\

    $ \Delta $ 
    & \textcolor{OliveGreen}{+0.2} 
    & \textcolor{OliveGreen}{+0.4} 
    & \textcolor{OliveGreen}{+0.9} 
    &\textcolor{OliveGreen}{+0.9}
    & \textcolor{OliveGreen}{+1.1} 
    & \textcolor{OliveGreen}{+0.8} 
    & \textcolor{red}{-1.1}  
    & \textcolor{OliveGreen}{+8.1} 
    & \textcolor{OliveGreen}{+1.8} 
    & \textcolor{OliveGreen}{+1.3} 
    &\textcolor{OliveGreen}{+3.7}
    & \textcolor{OliveGreen}{+2.3} 
    &\textcolor{OliveGreen}{+0.9} 
    &\textcolor{OliveGreen}{+2.6}\\ 

    \addlinespace[0.5em] 
    \cdashline{1-15}
    \addlinespace[0.5em] 

    RETFound \cite{zhou_foundation_2023}
    &81.2 
    &82.3 
    &77.7 
    &75.8 
    &82.1 
    &71.8
    &63.2 
    &63.0 
    &75.1  
    &42.5
    &70.1  
    &78.4 
    &76.9  
    &65.2\\

     \rowcolor[HTML]{C5E6F2} 
    + \methodname{} (Our)    
    &83.1
    &83.6
    &80.2
    &78.4
    &86.3
    &74.6
    &69.5
    &70.5
    &77.1
    &46.4
    &76.9
    &79.1
    &79.9
    &69.9 \\
    
    $ \Delta $ 
    & \textcolor{OliveGreen}{+0.9} 
    & \textcolor{OliveGreen}{+1.3} 
    & \textcolor{OliveGreen}{+2.5} 
    &\textcolor{OliveGreen}{+2.6}
    & \textcolor{OliveGreen}{+4.2} 
    & \textcolor{OliveGreen}{+2.8} 
    & \textcolor{OliveGreen}{+6.3}  
    & \textcolor{OliveGreen}{+7.5} 
    & \textcolor{OliveGreen}{+2.0} 
    & \textcolor{OliveGreen}{+3.9} 
    &\textcolor{OliveGreen}{+6.8}
    & \textcolor{OliveGreen}{+0.7} 
    &\textcolor{OliveGreen}{+3.0} 
    &\textcolor{OliveGreen}{+4.7}\\ 

    \addlinespace[0.5em] 
    \cdashline{1-15}
    \addlinespace[0.5em] 

    SwinV2$_{\text{tiny}}$ \cite{liu2022swin}
    &82.8 
    &83.4 
    &79.9 
    &78.1
    &85.9 
    &74.3
    &68.1 
    &67.6 
    &76.0  
    &44.1
    &76.2  
    & 81.5 
    &79.6  
    &68.9\\

    \rowcolor[HTML]{C5E6F2} 
    + \methodname{} (Our)     
    &83.0 
    &83.6 
    &80.2 
    &78.3 
    &86.3 
    &74.6 
    
    &69.5 
    &71.3 
    &77.1 
    &46.3 
    &77.1 
    &83.7 
    &79.9  
    &70.8 \\ 
    
    $ \Delta $ 
    & \textcolor{OliveGreen}{+0.2} 
    & \textcolor{OliveGreen}{+0.2} 
    & \textcolor{OliveGreen}{+0.3} 
    &\textcolor{OliveGreen}{+0.2}
    & \textcolor{OliveGreen}{+0.4} 
    & \textcolor{OliveGreen}{+0.3} 
    & \textcolor{OliveGreen}{+1.4}  
    & \textcolor{OliveGreen}{+3.7} 
    & \textcolor{OliveGreen}{+1.1} 
    & \textcolor{OliveGreen}{+2.2} 
    &\textcolor{OliveGreen}{+1.1}
    & \textcolor{OliveGreen}{+2.0} 
    &\textcolor{OliveGreen}{+0.3} 
    &\textcolor{OliveGreen}{+1.9}\\ 

    \addlinespace[0.5em] 
    \cdashline{1-15}
    \addlinespace[0.5em] 
    
    SwinV2$_{\text{large}}$ \cite{liu2022swin}
    &83.2 
    &83.6 
    &80.4 
    &79.0
    &87.2 
    &75.5
    &70.9 
    &73.5 
    &76.5  
    &48.2
    &77.4
    & 86.0 
    &80.5  
    &72.1\\

    \rowcolor[HTML]{C5E6F2} 
    + \methodname{} (Our)     
    & \textbf{83.2}
    & \textbf{83.6}
    & \textbf{80.4}
    & \textbf{79.1}
    & \textbf{87.3}
    & \textbf{75.8}
    & \textbf{71.2}
    & \textbf{74.5}
    & \textbf{77.1}
    & \textbf{48.2}
    & \textbf{77.6}
    & \textbf{86.2}
    & \textbf{80.7}
    & \textbf{72.5}\\

    $ \Delta $ 
    & \textcolor{OliveGreen}{+0.0} 
    & \textcolor{OliveGreen}{+0.0} 
    &\textcolor{OliveGreen}{+0.0}
    & \textcolor{OliveGreen}{+0.1} 
    & \textcolor{OliveGreen}{+0.1} 
    & \textcolor{OliveGreen}{+0.3} 
    & \textcolor{OliveGreen}{+0.3}  
    & \textcolor{OliveGreen}{+1.0} 
    & \textcolor{OliveGreen}{+0.6} 
    & \textcolor{OliveGreen}{+0.0} 
    &\textcolor{OliveGreen}{+0.2} 
    &\textcolor{OliveGreen}{+0.2}
    &\textcolor{OliveGreen}{+0.2} 
    &\textcolor{OliveGreen}{+0.4}\\

    \bottomrule
  \end{tabular}
  }

  \caption{\textbf{\methodname{} Results.} Quantitative comparison of \methodname{}-generated data integrated into DinoV2, RETFound, and SwinV2 across model sizes. Baselines are trained on datasets from \cref{sec:segmentation_datasets}. Evaluation spans \indomain{}, \neardomain{}, and \outdomain{} benchmarks, with average performance for \neardomain{} and \outdomain{}. Previous state-of-the-art performance (gray) reflects open-source inference or reported results. Performance is the average Dice score for artery and vein. $^{\dagger}$ indicates data leakage during training.}

  \label{tab:results}
\end{table*}

\subsection{Backbone Pretraining}
We investigate pretraining strategies to enhance segmentation performance, focusing on two key approaches: Masked Autoencoders (MAE) \cite{he2022masked} and Windowed Contrastive Learning (WCL) \cite{fan2023contrastive}. MAE facilitates robust representation learning by reconstructing masked inputs, effectively teaching the model to predict missing portions of an image. WCL, initially designed for depth estimation, employs contrastive learning on small image patches while maintaining local spatial relationships, making it particularly suitable for semantic segmentation tasks. Furthermore, we explore multi-objective pretraining \cite{li2023blip,yu_self-supervised_2024,fhima_tap-vl_2024}, by combining MAE and WCL to develop richer representations and improve downstream task performance. The dataset used for pretraining aligns with the one employed to train \methodname{}.

\subsection{Enhancing AV Segmentation with \methodname{}}
The synthetic images generated by \methodname{} serve as powerful data augmentation tools for vessel segmentation models. By preserving vascular structures while varying other characteristics (e.g., disc or lesions), these images enrich training datasets without requiring additional manual annotations.


Let a vessel segmentation model be denoted as $\mathcal{S}$, trained on real retinal images $x_{\text{orig}}$ with ground truth AV annotations $y$. The segmentation loss combines Dice loss and Binary Cross-Entropy (BCE) where $L^{\text{A}}$ and $L^{\text{V}}$ specifically represent the loss terms computed over artery and vein, respectively:
\begin{equation}
L_{\text{seg}} = 0.5 \cdot (L_{\text{Dice}}^{\text{A}} + L_{\text{BCE}}^{\text{A}}) + 0.5 \cdot (L_{\text{Dice}}^{\text{V}} + L_{\text{BCE}}^{\text{V}}).
\end{equation}

The total training objective includes supervised loss on real images and consistency loss on synthetic images:
\begin{equation}
L_{\text{total}} = L_{\text{seg}}(\mathcal{S}(x_{\text{orig}}), y) \;+\; \lambda \cdot L_{\text{seg}}(\mathcal{S}(x_{\text{gen}}), y),
\end{equation}
where $x_{\text{gen}}$ is a synthetic image sharing vascular structure with $x_{\text{orig}}$, and $\lambda>0$ balances contributions from real and synthetic data. This consistency regularization improves robustness across diverse imaging conditions, enhancing segmentation performance on unseen datasets.

\vspace{\baselineskip}

Additional implementation details, including hyperparameters and optimization strategies, are provided in the supplementary material.


\section{Experimental Setup}


We address data scarcity in retinal vessel segmentation by evaluating \methodname{}'s ability to generate controllable, realistic fundus images and improve AV segmentation performance. Key evaluations include image realism (\cref{sec:comparative_fid}), segmentation performance across backbones (\cref{sec:integration_backbone}), SOTA comparisons (\cref{sec:vs_sota}), and ablation studies (\cref{sec:abltation}).
We seek to address three key research questions:
\begin{itemize}
\item Can \methodname{} generate controllable, realistic retinal images?
\item Does usage of \methodname{}-generated data enhance our AV segmentation model?
\item How does our model perform compared to SOTA?
\end{itemize}

\subsection{Evaluation Metrics}

We evaluate the diffusion model's performance using the Fréchet Distance (FD), which compares the feature distributions of real and generated images. We compute it in the latent space of Inception-v3 (FID) \cite{heusel2017gans} and RETFound \cite{zhou_foundation_2023} (RET-FD), a foundation model pre-trained on 1.6 million retinal images. RETFound likely offers a more accurate representation of retinal image-specific features, while Inception-v3 enables a comparison with previous work. 

For AV segmentation, we use the Dice score to measure overlap between predicted and ground truth segmentations, averaged as  \((\text{Dice}_A + \text{Dice}_V)/2\). This is complemented by the Intersection over Union (IoU) and centerline Dice (clDice) \cite{cldice2021}, which emphasizes vessel centerlines. Both Dice and clDice metrics are employed in \methodname{} ablation studies, with additional IoU and clDice results provided in the supplementary material. Notably, clDice offers a more nuanced evaluation by balancing sensitivity to both thin and large vessels.




\subsection{Evaluation of Realism}

\label{sec:comparative_fid}
We compare the FID scores achieved by \methodname{} with those of prior works (Table~\ref{tab:fid_comparison}), using their publicly available models for image generation or reports their published results when the models were inaccessible. Notably, \methodname{} demonstrates superior performance by generating more realistic retinal fundus images, as evidenced by lower FID and RET-FD scores.

\begin{table}[b!]
  \centering
    \setlength{\tabcolsep}{3.5pt} 

  \begin{tabular}{l c c c c c c c}
    \toprule
    \textbf{Gen Model} &  \textbf{Conditioning} & \textbf{FID}$\downarrow$ & \textbf{RET-FD}$\downarrow$\\
    \midrule

     StyleGAN \cite{retinagan}&L&138.0&120.8\\

     StyleGAN2 \cite{mueller2024disentangling}&\small{Demographics}&98.1&116.0\\
    
    StyleGAN2 \cite{go2024generation}\textsuperscript{\textdagger}  &AV&122.8  &-\\
    Pix2PixHD 
    \cite{go2024generation}\textsuperscript{\textdagger}&AV  & 86.8 &- \\

    \rowcolor[HTML]{C5E6F2} 
    \methodname{} (Our)&AV + L + CD& \textbf{30.3} & \textbf{79.7} \\
    \bottomrule
  \end{tabular}

\caption{\textbf{Realism of Generated Images.} Lower FID and RET-FD on the DRTiD dataset indicate closer alignment with real data, reflecting realism. Notably, \methodname{} is able to generate controllable and more realistic retinal images. Models\textsuperscript{\textdagger} trained and evaluated on private data.}

\label{tab:fid_comparison}
\end{table}


\subsection{Integrating \methodname{} into Leading Backbones}
\label{sec:integration_backbone}
In Table~\ref{tab:results}, we present the performance of \methodname{}-generated data on the AV segmentation task, evaluated using various backbones: DinoV2${_\text{small}}$, RETFound, SwinV2${_\text{tiny}}$, and SwinV2$_{\text{large}}$. The results are reported across \indomain{}, \neardomain{}, and \outdomain{} test sets. For comparison, the first rows include previously published state-of-the-art results under similar settings (i.e., \indomain{}, \neardomain{}, and \outdomain{}), where available. 

\methodname{} consistently improves performance on \neardomain{}, and \outdomain{} test sets, demonstrating its backbone-agnostic advantages and its adaptability to in-domain and out-of-domain pretrained models. For example, integrating \methodname{} with RETFound yields performance improvements of 6.3\%, 7.5\%, and 6.8\% on AV-WIDE, IOSTAR, and TREND, respectively. Notably, even when applied to the top-performing backbone, SwinV2$_{\text{large}}$, \methodname{} provides further performance gains of 0.2\% on \neardomain{} and 0.4\% in \outdomain{} datasets.

\subsection{Segmentation performance vs SOTA}
\label{sec:vs_sota}

\begin{figure*}[ht] 
  \centering
    \includegraphics[width=\linewidth]{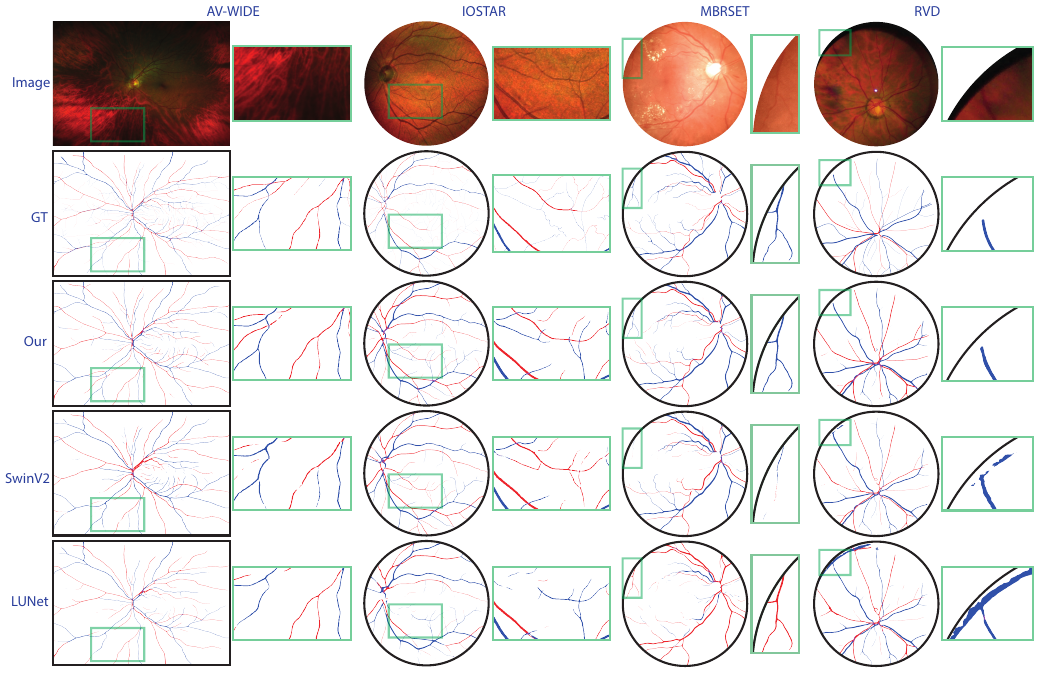}
\caption{\textbf{Qualitative Example on the Segmentation Downstream Task.} Comparing our model’s AV segmentation to a SwinV2$_\text{Large}$ \cite{liu2021swin} trained on the UZLF dataset and LUNet \cite{fhima2024lunet}, a SOTA model, showcasing its superior performance across fundus images from various datasets.}

    \label{fig:qualitative_segmentation}
\end{figure*}

SwinV2${_\text{large}}$, trained on our newly curated dataset and \methodname{}-generated data, surpasses previous state-of-the-art models across all \indomain{}, \neardomain{}, and \outdomain{} datasets, with the exception of RVD (Table \ref{tab:results}). As illustrated in Figure \ref{fig:qualitative_segmentation}, it demonstrates superior AV segmentation performance compared to SwinV2$_{\text{large}}$ trained solely on the UZLF dataset and LUNet, the best performing open-source model. Further quantitative and qualitative comparisons are included in the supplementary material. Moreover, a comprehensive analysis demonstrating the superiority of our model over previous state-of-the-art methods in estimating common vascular parameters is also provided in the supplementary material.

\section{Ablation studies}
\label{sec:abltation}
We analyze the effects of \methodname{}'s components, training datasets, and pretraining objectives using SwinV2$_{\text{tiny}}$ as the baseline and Dice score unless stated otherwise.

\textbf{Training Datasets:}  Starting with the UZLF dataset, we incrementally added our newly introduced datasets (Table~\ref{tab:ablation_dataset}). The \indomain{} test sets includes optic disc centered images, while \neardomain{} test sets mix optic disc and macula centered images. Adding macula-centered datasets GRAPE and MESSIDOR improved performance across \indomain{}, \neardomain{} and \outdomain{} test sets. Each dataset addition yielded incremental gains, with final improvements of +1.1\%, +4.1\%, and +8.3\% for \indomain{}, \neardomain{}, and \outdomain{}, respectively. 
\begin{table}[t!]
  \centering
    \resizebox{\linewidth}{!}{%

  \begin{tabular}{l c ccc}
    \toprule
    \textbf{Datasets} & \textbf{Size}& \textbf{\indomain{}} & \textbf{\neardomain{}} & \textbf{\outdomain} \\
    \midrule
    UZLF \cite{van_eijgen_leuven-haifa_2024}& 184 & 82.1 & 75.5 & 60.6 \\

    \addlinespace[0.5em] 
    \cdashline{1-5}
    \addlinespace[0.5em] 
    
    + GRAPE (Our$^{\dagger}$)& 81 & 82.6 & 78.1 & 65.2 \\
    + MESSIDOR (Our$^{\dagger}$)& 67 & 82.8 & 78.9 & 66.6 \\
    + ENRICH (Our$^{*}$)& 111 & 83.1 & 79.2 &67.0 \\
    + MAGRABIA (Our$^{\dagger}$)& 69 & 83.1 & 79.2 & 67.2\\
        \rowcolor[HTML]{C5E6F2} 
    + PAPILA (Our$^{\dagger}$)& 78 & \textbf{83.1} & \textbf{79.6}  & \textbf{68.9} \\

    $ \Delta $ & & \textcolor{OliveGreen}{+1.0} & \textcolor{OliveGreen}{+4.1} & \textcolor{OliveGreen}{+8.3} \\ 

    \bottomrule
  \end{tabular}
  }
\caption{\textbf{Impact of increasing the number of training datasets.} This table shows how adding newly introduced ($^{*}$) or annotated ($^{\dagger}$) datasets to the SwinV2$_{\text{tiny}}$ training pipeline impact performance.}

  \label{tab:ablation_dataset}
\end{table}

\begin{table}[t]
  \centering
  \setlength{\tabcolsep}{4pt} 
  \resizebox{\linewidth}{!}{%
  \begin{tabular}{ccc| cc cc cc }
    \toprule
    \multicolumn{2}{c}{\textbf{PT}}& \textbf{FT} &  \multicolumn{2}{c}{\textbf{\indomain{}}}&  \multicolumn{2}{c}{\textbf{\neardomain{}}} & \multicolumn{2}{c}{\textbf{\outdomain{}}} \\

    \cmidrule(lr){1-2} \cmidrule(lr){3-3} \cmidrule(lr){4-5} \cmidrule(lr){6-7} \cmidrule(lr){8-9}
\scriptsize MAE & \scriptsize WCL &\scriptsize Gen &\scriptsize Dice &\scriptsize clDice &\scriptsize Dice & \scriptsize clDice &\scriptsize Dice &\scriptsize clDice \\
    \midrule
    \xmark & \xmark & \xmark &83.1 &83.6& 79.6 & 80.7 & 68.9 & 68.8 \\

    \addlinespace[0.5em] 
    \cdashline{1-9}
    \addlinespace[0.5em] 

    \cmark & \xmark & \xmark &83.1&83.6  & 79.6 & 80.8 & 69.4 & 69.2 \\
    \xmark & \cmark & \xmark &83.2&83.6 & 79.7 & 80.8 & 69.2 & 69.1 \\
    \cmark & \cmark & \xmark &83.2&83.6& 79.6 & 80.8 & 69.4 & 69.3 \\

    \addlinespace[0.5em] 
    \cdashline{1-9}
    \addlinespace[0.5em] 

    \cmark & \cmark & AV&83.3& 83.7 & 79.9 & 81.1 & 70.4 & 70.5 \\

    \rowcolor[HTML]{C5E6F2} 

    \cmark & \cmark & AV + CD + L &\textbf{83.3}& \textbf{83.7} & \textbf{79.9} & \textbf{81.1} & \textbf{70.8} & \textbf{71.1} \\
    $\Delta $ & & & \textcolor{OliveGreen}{+0.2}&\textcolor{OliveGreen}{+0.1}& \textcolor{OliveGreen}{+0.3} & \textcolor{OliveGreen}{+0.4} & \textcolor{OliveGreen}{+1.9} & \textcolor{OliveGreen}{+2.3} \\ 

    \bottomrule
    
  \end{tabular}
  }
  \caption{\textbf{Pretraining Objective and Generation Method.} The top section shows baseline performance on our dataset, the middle highlights the impact of pretraining objectives, and the bottom examines AV conditioning versus AV + CD + L, with notable \outdomain{} improvements using AV + CD + L.}
  \label{tab:ablation_pretraining_components}
\end{table}



\textbf{Pretraining Objective:}
We evaluated how pretraining objectives (MAE, WCL, or both) influence our model's performance (see Table~\ref{tab:ablation_pretraining_components}). Adding MAE or WCL individually improved the \outdomain{} Dice score from 68.9\% to 69.2\% and 69.4\%, respectively, while combining them further increased clDice. These findings indicate that combining both strategies enhance model generalization.

\textbf{Conditioning on multiple \component s:} When learning a conditional distribution solely on AV, SwinV2$_\text{tiny+\methodname{}}$ achieved an average Dice score of 70.4\% on the \outdomain{} datasets. In contrast, conditioning on multiple \component s (AV, CD, and L) improved performance to 70.8\%. This highlights the advantage of leveraging a broader range of retinal fundus image features to enhance the learned distribution (see Table~\ref{tab:ablation_pretraining_components}).

\begin{figure}[H]
  \centering
  \includegraphics[width=\linewidth]{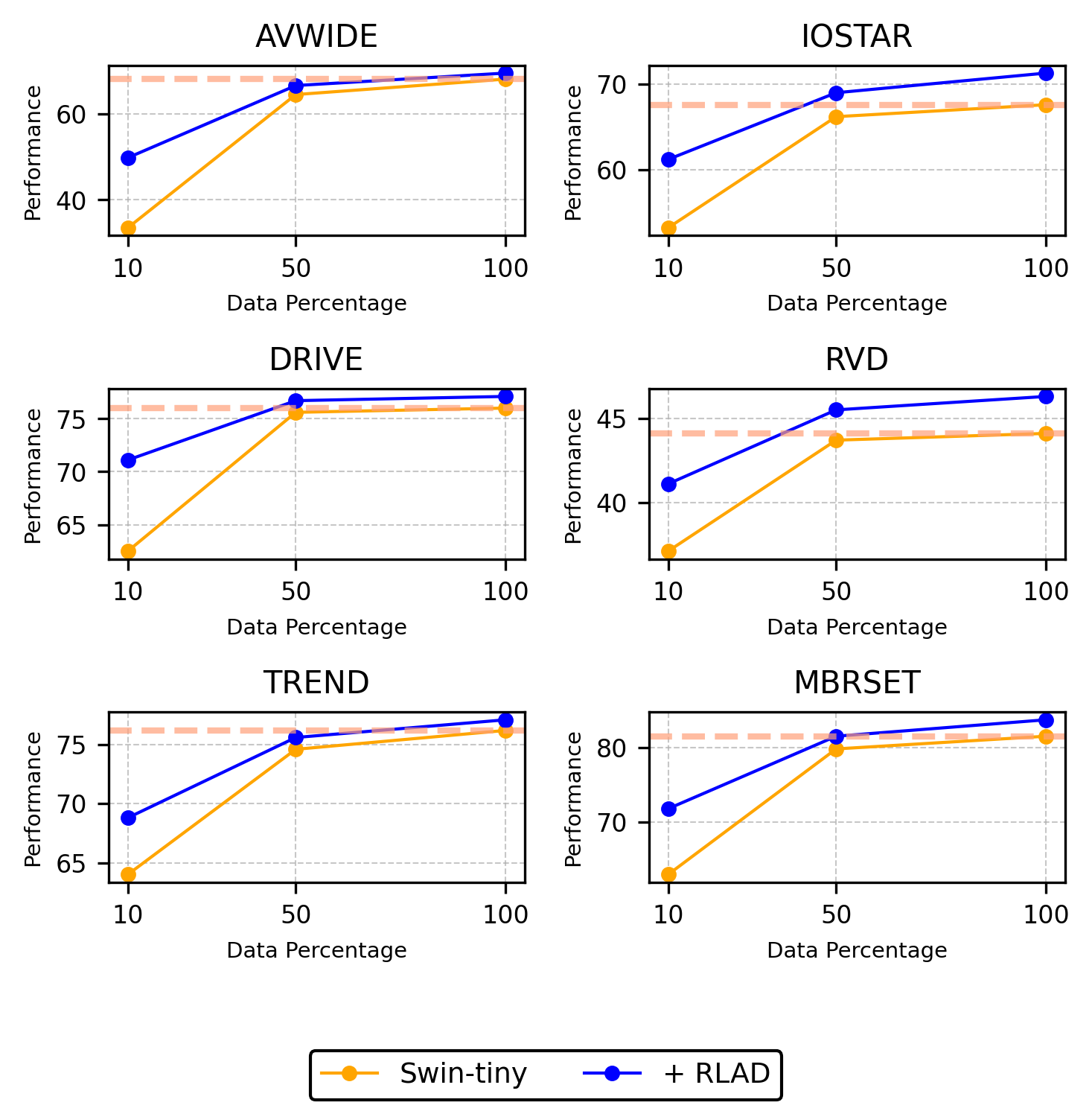}
  \caption{
    \textbf{\methodname{} Performance vs. Training Data Size.} The figure illustrates the learning curve of the SwinV2$_{\text{tiny}}$ \cite{liu2021swin} baseline on \outdomain{} datasets, demonstrating enhanced performance with \methodname{}-generated data. The data percentage reflects both real and generated samples, maintaining a 1:15 ratio (real:generated).
    }
  \label{fig:data_quantity}
\end{figure}

\begin{table}[b!]
  \centering
  \setlength{\tabcolsep}{4pt} 
  \resizebox{\linewidth}{!}{%
  \begin{tabular}{c| c c c c c c|c }
    \toprule
    \textbf{\# Gen} & \textbf{AV-WIDE} & \textbf{IOSTAR}& \textbf{DRIVE} & \textbf{RVD} & \textbf{TREND}  & \textbf{MBRSET}& \textbf{\outdomain{}} \\

    \midrule

    0.5K & 69.2  & 69.9 &77.2&45.8& 76.9 & 75.9 & 70.4\\
    
    1.5K & 69.5  & 70.5 & 77.1&46.4&76.9 & 76.0&  70.6\\
    
    \rowcolor[HTML]{C5E6F2} 7.2K & 69.5  & 71.3 & 77.1 &46.3&  77.1 & 76.2 & \textbf{70.8}\\

    \bottomrule
  \end{tabular}
  }
\caption{\textbf{Quantity of Generated Data.} We evaluate the impact of increasing \methodname{}'s generated data on performance, reporting Dice scores for each \outdomain{} dataset and their average performance.}

\label{tab:ablation_method_data}
\end{table}

\textbf{Varying Generated Data Quantity:} We explored the impact of varying amounts of \methodname{}-generated samples: 0.5K (1 per real image), 1.5K (3 per real image), and 7.2K (15 per real image). Increasing generated samples improved the average \outdomain{} Dice (Table~\ref{tab:ablation_method_data}) and clDice (see supplementary material).


\textbf{Performance Gains of \methodname{} Relative to Dataset Size:} Figure~\ref{fig:data_quantity} shows learning curves on \outdomain{} datasets for SwinV2$_{\text{tiny}}$ trained with and without \methodname{} synthetic data. Incorporating \methodname{}-generated data consistently improves performance across all datasets. For IOSTAR, RVD, DRIVE, and MBRSET, the model trained with synthetic data outperformed the baseline while using less than 50\% of the baseline’s training data. The largest gains occurred in data-scarce scenarios, highlighting \methodname{}'s effectiveness in enhancing performance.



\section{Conclusion}
This work presents \methodname{}, a novel diffusion-based framework designed to generate realistic and controllable retinal fundus images by conditioning on multiple \component s extracted from real-world data. 
Beyond image generation, \methodname{} proves to be a valuable tool for advancing downstream tasks. By incorporating the synthetic data generated by \methodname{}, we significantly enhance the training datasets for AV segmentation tasks, resulting in notable performance improvements across various visual backbones. This capability is particularly impactful in data-scarce scenarios, where access to comprehensive datasets is limited.
Our findings highlight the potential of \methodname{} to drive innovation in medical imaging applications and improve segmentation outcomes. Future research could explore its application to other imaging modalities and investigate optimization strategies to further enhance its adaptability and scalability.

{
    \small
    \bibliographystyle{ieeenat_fullname}
    \bibliography{main.bib}
}

\clearpage
\setcounter{page}{1}
\maketitlesupplementary
\appendix

\begin{minipage}{\textwidth}
  \vspace{0.5em} 
  \listappendices 
\end{minipage}

\clearpage

\appsection{Datasets}
\label{app_dataset}
For our experiments, we utilized two distinct dataset combinations to support both the training and evaluation phases of our methodology.

The dataset tables provide a comprehensive summary of the key characteristics of each dataset, including the number of samples, the primary pathology—glaucoma (G), diabetic retinopathy (DR), age-related macular degeneration (AMD), or multiple different diseases (Multiple)—the imaging center, which is either disc (D) or macula (M), field of view (FOV), geographic region, and image resolution.

\appsubsection{Diffusion and Pretraining Datasets}

The first combination involved non-annotated datasets used for training the \methodname{} model and pretraining segmentation models, as summarized in Table \ref{tab:pretraining_data_table}. 

\appsubsection{Segmentation Datasets}
The second combination comprised AV-annotated datasets, which were employed for training segmentation models on downstream tasks (Table \ref{tab:downstream_data_table}) and for evaluating their performance (Table \ref{tab:benchmark_data_table}). Furthermore, the AV segmentation datasets released within \datasetname{} are summarized in Table \ref{tab:reyia}. Datasets annotated specifically for this study are marked with $^{\dagger}$, while those introduced and annotated as part of this work are marked with $^{*}$.

\setlength{\tabcolsep}{4pt}
\begin{table}[H]
    \centering
    \small
    \setlength{\tabcolsep}{5pt}
    
    \resizebox{0.48 \textwidth}{!}{
    \begin{tabular}{l c c c c}
            \toprule
       \textbf{Dataset} & \textbf{\# Samples} & \textbf{Image Center} & \textbf{FOV ($^\circ$)} \\

            \midrule

            GRAPE$^{\dagger}$ \cite{huang2023grape}& 81 & M & 50 \\
            MESSIDOR$^{\dagger}$ \cite{messidor2011}& 67 & M & 45  \\
            PAPILA$^{\dagger}$ \cite{kovalyk2022papila}& 78 & D & 30 \\
            MAGHREBIA$^{\dagger}$ \cite{Almazroa2018RIGA}& 69 & M, D & 30 \\
            ENRICH$^{*}$ & 111 & D & 45 \\

            FIVES$^{\dagger}$ \cite{jin2022fives}& 75 & M & 45 \\

            AV-WIDE$^{\dagger}$ \cite{estrada2015retinal}& 27  & D & Ultra wide \\
         
          TREND $^{\dagger}$ \cite{popovic_trend_2021}& 48  & M & 30 \\
          
           MBRSET$^{\dagger}$ \cite{wu2024mbrset}& 30 & M & 30 \\
           
            \bottomrule
        \end{tabular}
    }
  \caption{\textbf{List of the dataset included in the \datasetname{} collection} released with this work. Datasets marked with $^{\dagger}$ were annotated specifically for this work, and those marked with $^{*}$ were both introduced and annotated here.}
  \label{tab:reyia}

\end{table}
\setlength{\tabcolsep}{1.4pt}

\appsection{Training Hyperparameters}
All experiments were conducted on 4 Nvidia A100 (40G) GPUs using bfloat16 precision. In each training the AdamW optimizer \cite{DBLP:conf/iclr/LoshchilovH19} and the Cosine Annealing scheduler \cite{cosine} were uniformly applied. Beyond these constants, each training was characterized by its own distinct set of hyperparameters. 

\textbf{\methodname{} Training:} comprised 84,000 training steps, with a learning rate \(1e-4\) and and a batch size of $12$. 

\textbf{Segmentation Models Pretraining:} comprised 1 training epoch, with a learning rate \(1.5e-4\) and and a batch size of $128$.  

\textbf{Segmentation Models Finetuning:} comprised 200 training epochs, with a learning rate \(4e-4\). Other hyperparameters varied based on the backbone and are described in Table \ref{tab:implementation_table_stage3}.
\setlength{\tabcolsep}{4pt}
\begin{table}[H]
    \centering
    \small
    \setlength{\tabcolsep}{5pt}
    
    \resizebox{0.49 \textwidth}{!}{
        \begin{tabular}{lcccccccc}
            \toprule
            \textbf{Backbone} & \textbf{\# Epochs} & \textbf{\# Batch Size}&\textbf{Learning Rate} & \textbf{$\boldsymbol{\lambda}$} \\

            \midrule

            DinoV2$_{\text{small}}$ \cite{oquab2023dinov2}  &$200$&12&$4e-4$&$1.0$\\

            RETFound \cite{zhou_foundation_2023} &$200$& 12 &$4e-4$&$0.1$\\
            
            SwinV2$_{\text{tiny}}$ \cite{liu2022swin}&12&$200$&$4e-4$&$0.1$\\

            SwinV2$_{\text{large}}$ \cite{liu2022swin}&2&$200$&$4e-4$&$0.1$\\
           
            \bottomrule
        \end{tabular}
    }
    \caption{\textbf{Hyperparameters for the segmentation downstream task finetuning.}}
        \label{tab:implementation_table_stage3}

\end{table}
\setlength{\tabcolsep}{1.4pt}

\appsection{Additional Quantitative Results}
\label{appendix:additionalResults}

In addition to the metrics reported in the main paper, we report Intersection over Union (IoU) and centerline Dice score (clDice) for SwinV2$_\text{Large + RLAD}$ versus the open -souce models. IoU measures the ratio of the intersection to the union of the predicted and ground truth segmentation masks, providing an additional evaluation of segmentation performance. The IoU is computed separately for arteries (A) and veins (V), and we report the average IoU across both classes \((\text{IoU}_A + \text{IoU}_V)/2\). This metric complements the Dice score by offering a stricter evaluation of overlap, particularly for challenging cases with smaller or less distinct structures.
Table \ref{tab:additional_results} shows that our model outperform all open-source baseline for both clDice and IoU across all datasets, except the DRIVE where VascX \cite{quiros_vascx_2024} get higher IoU performance.

\appsection{Additional Qualitative Results}
\label{appendix:additionalQualiResults}

\begin{figure}[htb] 
  \centering
    \includegraphics[width=\linewidth]{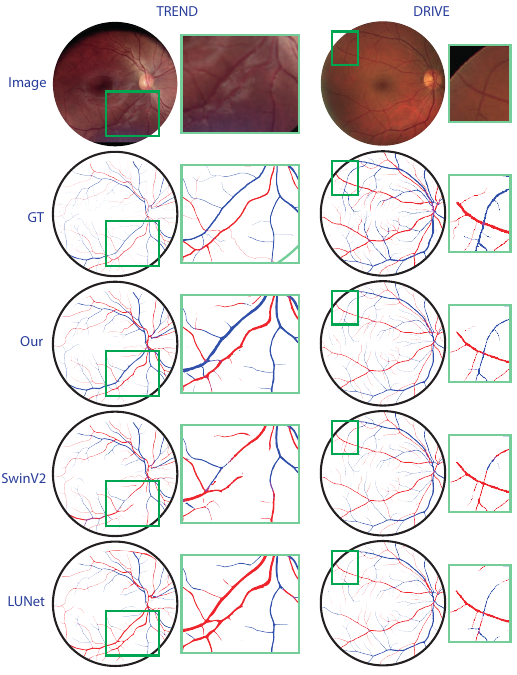}
\caption{\textbf{Qualitative Example on the Segmentation Downstream Task.} Comparing our model’s AV segmentation to a SwinV2$_\text{large}$ \cite{liu2022swin} trained on the UZLF dataset and SOTA model LUNet \cite{fhima2024lunet}, showcasing its superior performance across diverse fundus images.}
    \label{fig:qualitative_segmentation_additional}
\end{figure}

In Figure \ref{fig:qualitative_segmentation_additional}, we display some additional qualitative examples of our model compared to a SwinV2$\text{large}$ baseline and a SOTA open-source model LUNet. We can see that our model more accurately segment the blood vessels of the DRIVE and TREND datasets.

\appsection{Additional Ablation Results}
\label{app:additional_ablation}
Additional ablation results on the impact of the scale of the generated samples using clDice score are shown in Table \ref{tab:cldiceablation_method_data}. It shows that using more \methodname{}-generated samples also increased the average \outdomain{} performance for the clDice score.
\begin{table}[H]
  \centering
  \setlength{\tabcolsep}{4pt} 
  \resizebox{\linewidth}{!}{%
  \begin{tabular}{c| c c c c c c|c }
    \toprule
    \textbf{\# Gen} & \textbf{AV-WIDE} & \textbf{IOSTAR}& \textbf{DRIVE} & \textbf{RVD} & \textbf{TREND}  & \textbf{MBRSET}& \textbf{\outdomain{}} \\

    \midrule

    0.5K & 70.9  & 67.5 &79.6&46.2& 75.9 & 83.7 & 70.6\\
    
    1.5K & 70.9 & 68.2 & 79.6 & 46.8 & 76.0 & 84.0 & 70.9\\
    
    \rowcolor[HTML]{C5E6F2} 7.2K &70.9 & 69.0 & 79.7 & 46.6 & 76.2 & 84.2 & \textbf{71.1}\\

    \bottomrule
  \end{tabular}
  }
\caption{\textbf{Quantity of Generated Data.} We evaluate the impact of increasing \methodname{}'s generated data on performance, reporting clDice scores for each \outdomain{} dataset and their average performance.}
\label{tab:cldiceablation_method_data}

\end{table}

\begin{figure*}[b] 
  \vspace{-0.6cm}
  \centering
  \includegraphics[width=\linewidth]{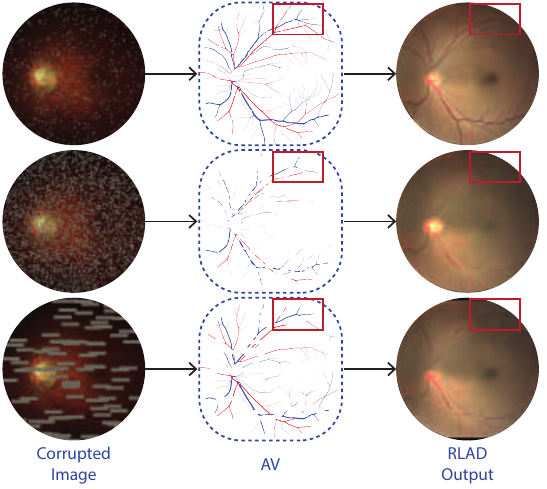} 
  \vspace{-0.2cm}
  \caption{Impact of the layout extractor.}
    \label{fig:feature_atrous}
  \vspace{-0.3cm}
\end{figure*}

\appsection{Impact of the Layout Extractor} \methodname{} is trained on an approximation of the layout extracted by a deep learning model, rather than relying on a ground truth conditioning. This enables \methodname{} to learn a distribution $p_\theta(x_{t-1}|\widehat{\text{layout}},x_t)$ instead of $p_\theta(x_{t-1}|\text{layout},x_t)$, allowing the model to adapt to noisy conditioning. Consequently, \methodname{} exhibits a degree of robustness to the errors typically made by the layout extractor. Figure \ref{fig:feature_atrous} illustrates this with intentionally corrupted images, generated by applying a random masking strategy. While the extracted blood vessels are impacted by the corruption, the final images generated by \methodname{} remain relatively unaffected, provided the density of the masks is limited. This robustness aligns with the known limitations of current retinal blood vessel segmentation models. Thus, we assume that the performance of the Layout Extractor remains a relatively unimportant factor (for small performance differences), given that its limitations will be mitigated by the diffusion model.

\appsection{Vascular Parameters Estimation}
\label{appendix:vbms}
Vascular parameters were estimated using the PVBM toolbox \cite{fhima2022pvbm}, including area (Area), tortuosity indices (TI, TOR), length (LEN), branching angles (BA), key vascular points (SPoints, EPoints, BPoints), fractal dimensions (D0, D1, D2, SL), and retinal metrics (CRAE/CRVE, AVR). Parameters were evaluated on \outdomain{} datasets by computing Pearson correlations between ground-truth and estimated values, with final scores representing averages across datasets and vascular structures (arteries/veins).
\begin{table}[htb]
  \centering
    \setlength{\tabcolsep}{3pt} 
  \resizebox{0.9\linewidth}{!}{%
  \begin{tabular}{l | c c c c c}
    \toprule
     
      \small{\textbf{Vascular}} & \multirow{2}{*}{\small{Little W-Net}} & \multirow{2}{*}{\small{Automorph}} & \multirow{2}{*}{\small{VascX}} & \multirow{2}{*}{\small{LUNet}} &  \multirow{2}{*}{\small{Our}} \\
    \small{\textbf{Parameters}} \\
    \midrule
     Area& 55.7&\textbf{73.2}&69.9&61.3&71.4\\
     TI& 46.3&61.3&62.9&59.3&\textbf{71.7}\\ 
     TOR&45.6&53.9&61.7&60.6&\textbf{68.8}\\
     LEN& 56.8&69.3&68.9&68.5&\textbf{75.5}\\
     BA& 24.3& 45.5&44.8&38.6&\textbf{51.5}\\
     SPoints& 41.6&56.4&56.7&55.4&\textbf{62.2}\\
     EPoints&53.7&70.1&71.3&68.3&\textbf{77.7}\\
     BPoints&39.4&55.3&55.8&53.1&\textbf{65.2}\\
     D0& 56.0&59.8&65.3&61.6&\textbf{69.0}\\
     D1& 60.9&68.3&73.2&72.7&\textbf{80.7}\\
     D2& 48.7&54.1&58.7&60.8&\textbf{70.0}\\
     SL& 48.8&53.8&54.3&59.0&\textbf{63.6}\\
     CRE$_\text{H}$& 55.1&66.0&66.8&69.9&\textbf{75.8}\\
     CRE$_\text{K}$& 52.1&65.5&62.1&67.4&\textbf{75.0}\\
     AVR$_\text{H}$& 66.7&74.3&78.9&78.2&\textbf{81.0}\\
     AVR$_\text{K}$& 31.4&41.9&44.1&47.4&\textbf{52.9}\\
     \cdashline{1-6}

     Average& 48.9& 60.5 & 61.4 & 62.2 & \textbf{69.5}\\

    \bottomrule
  \end{tabular}
  }
  \caption{\textbf{\methodname{} Vascular Parameters Results}. Quantitative comparison of SwinV2$_\text{Large + \methodname{}}$ (Our) versus open-source models. Performance is reported as the average Pearson correlation coefficient in estimating vascular parameters across \outdomain{} datasets.}
  
  \label{tab:vbms}
\end{table}

\begin{table*}[hb]
  \centering
  \small
  \setlength{\tabcolsep}{4pt} 
  \makebox[\textwidth][c]{%
    \begin{tabular}{l c c c c c c}
      \toprule
       \textbf{Dataset} & \textbf{\# Samples} & \textbf{Primary Pathology} & \textbf{Image Center} & \textbf{FOV ($^\circ$)} & \textbf{Region} & \textbf{Resolution (px)} \\
      \midrule
           UZLF \cite{van_eijgen_leuven-haifa_2024}& 184 & G & D & 30 & Belgium & 1444×1444 \\
            GRAPE \cite{huang2023grape}& 81 & G & M & 50 & China & 1444×1444 \\
           MESSIDOR \cite{messidor2011}& 67 & DR & M & 45 & France & 1444×1444 \\
            PAPILA \cite{kovalyk2022papila}& 78 & G & D & 30 & Spain & 1444×1444 \\
            MAGHREBIA \cite{Almazroa2018RIGA}& 69 & -- & M, D & 30 & Maghreb & 1444×1444 \\
           ENRICH & 111 & G & D & 45 & Belgium & 1958×2196 \\
           1000images \cite{cen_automatic_2021}& 973 & Multiple & D & 30 & China & 3000x3152\\
           DDR \cite{DDR2019}& 12 519 & DR & M & 45 & China & 1728x2592\\
           EYEPACS \cite{EYEPACS} &88 702& DR& M & 45 & United States & VAR\\
           G1020 \cite{G1020}& 1020 & G & M & 45 & Germany& 2423x3004\\
           IDRID \cite{IDRID2018}& 516 & DR & M & 50 & India & 2848x4288\\
           ODIR \cite{ODIR2019}& 8000& Multiple & M & 45 & China & 1296x1936\\
      \bottomrule
    \end{tabular}
  }
  \caption{\textbf{Summary of Datasets Used for Pretraining and \methodname{} Training.} This table lists the datasets used for pretraining segmentation models and training the \methodname{} framework. Key attributes include the number of samples, primary pathologies, imaging center type, field of view (FOV), geographic region, and resolution. 
  }
  \label{tab:pretraining_data_table}
\end{table*}

\begin{table*}[hb]
  \centering
  \small
  \setlength{\tabcolsep}{4pt} 
  \makebox[\textwidth][c]{%
    \begin{tabular}{l c c c c c c}
      \toprule
       \textbf{Dataset} & \textbf{\# Samples} & \textbf{Primary Pathology} & \textbf{Image Center} & \textbf{FOV ($^\circ$)} & \textbf{Region} & \textbf{Resolution (px)} \\
      \midrule
           UZLF \cite{van_eijgen_leuven-haifa_2024}& 184 & G & D & 30 & Belgium & 1444×1444 \\
            GRAPE$^{\dagger}$ \cite{huang2023grape}& 81 & G & M & 50 & China & 1444×1444 \\
            MESSIDOR$^{\dagger}$ \cite{messidor2011}& 67 & DR & M & 45 & France & 1444×1444 \\
            PAPILA$^{\dagger}$ \cite{kovalyk2022papila}& 78 & G & D & 30 & Spain & 1444×1444 \\
            MAGHREBIA$^{\dagger}$ \cite{Almazroa2018RIGA}& 69 & -- & M, D & 30 & Maghreb & 1444×1444 \\
            ENRICH$^{*}$ & 111 & G & D & 45 & Belgium & 1958×2196 \\

      \bottomrule
    \end{tabular}
  }
  \caption{\textbf{Summary of Datasets Used for Downstream Segmentation Training.} This table lists the annotated datasets used for training segmentation models in downstream tasks. Attributes include the number of samples, primary pathologies, imaging center type, field of view (FOV), geographic region, and resolution. Datasets marked with $^{\dagger}$ were annotated specifically for this work, and those marked with $^{*}$ were both introduced and annotated here.}
  \label{tab:downstream_data_table}
\end{table*}

\begin{table*}[htbp]
  \centering
  \small
  \setlength{\tabcolsep}{4pt} 
  \makebox[\textwidth][c]{%
    \begin{tabular}{l c c c c c c c c}
      \toprule
       & \textbf{Dataset} & \textbf{\# Samples} & \textbf{Primary Pathology} & \textbf{Image Center} & \textbf{FOV ($^\circ$)} & \textbf{Region} & \textbf{Resolution (px)} \\
      \midrule
      \multirow{2}{*}{\textbf{\indomain{}}} 
          & UZLF-test \cite{van_eijgen_leuven-haifa_2024}& 56 & G & D & 30 & Belgium & 1444×1444 \\
          & LES-AV \cite{orlando2018towards}& 20 & G & D & 30 & Belgium & 1444×1444 \\
          
      \addlinespace[0.5em]
      \cdashline{1-8}
      \addlinespace[0.5em]

      \multirow{5}{*}{\textbf{\neardomain{}}}
         & HRF \cite{rombach2022high}& 45 & DR, G & M & 45 & Germany & 2336×3504 \\
         & INSPIRE \cite{Niemeijer2011Automated,fhima2024lunet}& 15 & -- & D & 30 & USA & 1444×1444 \\
          & FIVES$^{\dagger}$ \cite{jin2022fives}& 75 & DR, G, AMD & M & 45 & China & 1444×1444 \\
          & UNAF \cite{Benitez2021Dataset,fhima2024lunet}& 15 & DR & D & 30 & Paraguay & 2056×2124 \\

      \addlinespace[0.5em]
      \cdashline{1-8} 
      \addlinespace[0.5em]

      \multirow{7}{*}{\textbf{\outdomain{}}}
         & AV-WIDE$^{\dagger}$ \cite{estrada2015retinal}& 27 & -- & D & Ultra wide & USA & 829×1531 \\
          & IOSTAR \cite{iostar} & 30 & -- & M & 45 & Netherlands & 1024×1024 \\
          & DRIVE \cite{staal_ridge-based_2004,hu_automated_2013}& 40 & DR & M & 45 & Netherlands & 584×565 \\
          & RVD \cite{khan2024rvd}& 1270 & -- & VAR & 30 & -- & 1800x1800 \\
          & TREND $^{\dagger}$ \cite{popovic_trend_2021}& 48 & H & M & 30 & Montenegro & 2560x2560 \\
          & MBRSET$^{\dagger}$ \cite{wu2024mbrset}& 30 & DR, G, AMD & M & 30 & Brazil & 1444×1444 \\

      \bottomrule
    \end{tabular}
  }
  \caption{\textbf{Summary of Datasets Used for Segmentation Benchmark Evaluation.} This table categorizes datasets into in-domain (\indomain{}), near-domain (\neardomain{}), and out-of-domain (\outdomain{}) groups for evaluating segmentation performance. Attributes include the number of samples, primary pathologies, imaging center type, field of view (FOV), geographic region, and resolution. Datasets marked with $^{\dagger}$ were annotated specifically for this work, and those marked with $^{*}$ were both introduced and annotated here.}
  \label{tab:benchmark_data_table}
\end{table*}

\begin{table*}[t!]
  \centering
    \setlength{\tabcolsep}{3pt} 
  \resizebox{\linewidth}{!}{%
\begin{tabular}{l cc |  c c | c c | c c | c c | c c | c c | c c | c c | c c}
    \toprule
     \multirow{2}{*}{\textbf{Backbone}} & \multicolumn{8}{c}{\textbf{\neardomain{}}} & \multicolumn{12}{c}{\textbf{\outdomain{}}}\\ 
     & \multicolumn{2}{c}{HRF} & \multicolumn{2}{c}{INSPIRE} & \multicolumn{2}{c}{FIVES} & \multicolumn{2}{c}{UNAF} & \multicolumn{2}{c}{AV-WIDE} &\multicolumn{2}{c}{IOSTAR} & \multicolumn{2}{c}{DRIVE} & \multicolumn{2}{c}{RVD} & \multicolumn{2}{c}{TREND} & \multicolumn{2}{c}{MBRSET} \\
    \cmidrule(lr){2-9} \cmidrule(lr){10-21}

    & clDice & IoU &clDice & IoU &clDice & IoU &clDice & IoU &clDice & IoU &clDice & IoU &clDice & IoU &clDice & IoU &clDice & IoU &clDice & IoU  \\

    Little W-Net \cite{galdran2022state}
    &53.3& 41.5 
    &70.7& 55.6 
    &71.9& 59.0 
    &68.5& 52.5 

    &41.1& 28.1 
    &26.6& 19.3 
    &59.7& 44.4 
    &32.1& 22.2 
    &51.9& 36.9 
    &35.2& 34.6 
    \\

    Automorph \cite{zhou_automorph_2022}
    & 76.7$^\dagger$ & 63.3$^\dagger$ 
    &71.5& 55.3 
    &72.1& 57.9 
    &66.3& 49.9 

    &49.9& 33.9 
    &52.3& 38.4 
    & 77.3$^\dagger$ & 64.1$^\dagger$ 
    & 31.6 & 22.6 
    &65.3& 50.4 
    &62.0& 47.8 
    \\

    VascX \cite{quiros_vascx_2024}
    & 73.1&61.0 
    &75.3&60.0 
    &79.1&67.6 
    & 74.3&57.9 

    &49.7& 34.1 
    & 49.0& 35.6 
    &75.9& \textbf{63.5} 
    &39.7&28.1 
    & 69.6& 56.4 
    &73.4& 58.3 
    \\

    LUNet \cite{fhima2024lunet}
    &72.8&58.1 
    &76.4&64.9 
    &82.6&75.9 
    &76.7&59.5 

    &65.5& 53.4 
    &52.1& 40.2 
    &71.3& 55.4 
    &36.1& 22.4 
    &69.6& 55.9 
    &64.0& 48.0 
    \\

    \rowcolor[HTML]{C5E6F2} SwinV2$_\text{Large + \methodname{}}$ (Our)
    & \textbf{81.1}&\textbf{67.5} 
    & \textbf{83.0}&\textbf{65.5} 
    & \textbf{86.9}&\textbf{77.7} 
    & \textbf{78.3}&\textbf{61.4}  

    & \textbf{73.2} &\textbf{55.7} 
    & \textbf{73.0}& \textbf{59.8} 
    & \textbf{80.3}& 62.9 
    & \textbf{49.1}& \textbf{33.0} 
    & \textbf{77.9}& \textbf{63.8} 
    & \textbf{86.8}& \textbf{76.0} 
    \\

    \bottomrule
  \end{tabular}
  }
  \caption{\textbf{Additional \methodname{} Results}. Quantitative comparison of SwinV2$_\text{Large + \methodname{}}$ versus open source models. Performance is the average clDice/IoU for artery and vein. $^{\dagger}$ indicates data leakage during training.}
  \label{tab:additional_results}
\end{table*}

\end{document}